# A Hierarchical Structured Self-Attentive Model for Extractive Document Summarization (HSSAS)

KAMAL AL-SABAHI, ZHANG ZUPING, AND MOHAMMED NADHER

School of Information Science and Engineering, Central South University, Changsha, China

k.alsabahi@csu.edu.cn; zpzhang@csu.edu.cn

**ABSTRACT** The recent advance in neural network architecture and training algorithms has shown the effectiveness of representation learning. The neural-network-based models generate better representation than the traditional ones. They have the ability to automatically learn the distributed representation for sentences and documents. To this end, we proposed a novel model that addresses several issues that are not adequately modeled by the previously proposed models, such as the memory problem and incorporating the knowledge of document structure. Our model uses a hierarchical structured self-attention mechanism to create the sentence and document embeddings. This architecture mirrors the hierarchical structure of the document and in turn enables us to obtain better feature representation. The attention mechanism provides extra source of information to guide the summary extraction. The new model treated the summarization task as a classification problem in which the model computes the respective probabilities of sentence–summary membership. The model predictions are broken up by several features such as information content, salience, novelty, and positional representation. The proposed model was evaluated on two well-known datasets, the CNN/Daily Mail and DUC 2002. The experimental results show that our model outperforms the current extractive state of the art by a considerable margin.

## I. INTRODUCTION

Text summarization is one of the most active research in natural language processing. It is an optimal way to tackle the problem of information overload by reducing the size of long document(s) into a few sentences or paragraphs. The popularity of handheld devices, such as smart phone, makes document summarization very urgent in the face of tiny screens and limited bandwidth [1]. It can also serve as a reading comprehension test for machines. To generate a good summary, it is important for a machine learning model to be able to understand the document(s) and distill the important information from it. These tasks are highly challenging for computers, especially as the size of a document increases. Even though the most sophisticated search engines are empowered by advanced information retrieval techniques, they lack the ability to synthesize information from multiple sources and to give users a concise yet informative response. Moreover, there is a need for tools that provide timely access to, and digest of, various sources. These concerns have sparked a great interest in the development of automatic document summarization systems. Traditional Text Summarization approaches typically rely on sophisticated feature engineering that based mostly on the statistical properties of the document being summarized. In short, these systems are complex, and a lot of engineering effort goes into building them. On the top of that, those methods mostly fail to produce a good document representation and a good summary as a result. End-to-end learning models are interesting to try as they demonstrate good results in other areas, like speech recognition, language translation, image recognition, and even question-answering. Recently, neural network-based summarization approaches draw much attention; several models have been proposed and their applications to the news corpus were demonstrated, as in [2] and [3].

There are two common types of neural text summarization, extractive and abstractive. Extractive summarization models automatically determine and subsequently concatenate relevant sentences from a document to create its summary while preserving its original information content. Such models are common and widely used for practical applications [2], [3]. A fundamental requirement in any extractive summarization model is to identify the salient sentences that represent the key information mentioned in [4]. In contrast, abstractive text summarization techniques attempt to build an internal semantic representation of the original text and then create a summary closer to a human-generated one. The state-of-the-art abstractive models are still quite weak, so most of the previous work has focused on the extractive summarization [5].

Despite their popularity, neural networks still have some issues while applied to document summarization task. These methods lack the latent topic structure of contents. Hence the summary lies only on vector space that can hardly capture multi-topical content [6]. Another issue is that the most common architectures for Neural Text Summarization are variants of recurrent neural networks (RNN) such as Gated recurrent unit (GRU) and Long short-term memory (LSTM). These models have, in theory, the power to 'remember' past decisions inside a fixed-size state vector; however, in practice, this is often not the case. Moreover, carrying the semantics along all-time steps of a recurrent model is relatively hard and not necessary [7]. In this work, we use a hierarchical structured self-attention mechanism to tackle the problem. In which, a weighted average of all the previous states will be used as an extra input to the function that computes the next state. This gives the model the ability to attend to a state produced several time steps earlier, so the latest state does not have to store all the information [8].

The contribution of this paper is proposing a general neural network-based approach for summarization that extracts sentences from a document by treating the summarization problem as a classification task. The model computes the score for each sentence towards its summary membership by modeling abstract features such as content richness, salience with respect to the document, redundancy with respect to the summary and the positional representation. The proposed model has two improvements that enhance the summarization effectiveness and accuracy: (i) it has a hierarchical structure that reflects the hierarchical structure of documents; (ii) while building the document representation, two levels of self-attention mechanism are applied at word-level and sentence-level. This enables the model to attend differentially to more and less important content.

In this paper, two interesting questions are arising: (1) how to mirror the hierarchical structure of the document to improve the embedding representation of sentence and document that can help discovering the coherent semantic of the document; (2) how to extract the most important sentences from the document to form a desired summary [6].

The key difference between this work and the previous ones is that our model uses a hierarchical structured self-attention mechanism to create sentence and document embeddings. The attention serves two benefits: not only does it often result in better performance, but it also provides insight into which words and sentences contribute to the document representation and to the selection process of the summary sentences as well. To evaluate the performance of our model in comparison to other common state-of-the-art architectures, two well-known datasets are used, the CNN/Daily Mail, and DUC 2002. The proposed model outperforms the current state-of-the-art models by a considerable margin.

The rest of the paper is organized as follows. In section 2, the proposed approach for summarizing documents is presented in details. Section 3 describes the experiments and the results. The related work is briefly described in section 4. Finally, we discuss the results and conclude in section 5.

## II. THE PROPOSED MODEL

Recurrent Neural Network variants, such as LSTM, have been used widely in text summarization problem. To prepare the text tokens to be used as an input to these networks, word embeddings, as language models [9], [10], are used to convert language tokens to vectors. Moreover, attention mechanisms [11] make these models more effective and scalable, allowing them to focus on some past parts of the input sequence while making the final decision or generating the output.

*Definition 1: Given a document D consisting of a sequence of sentences $(s_1, s_2 \ldots, s_n)$ and a set of words $(w_1, w_2, \ldots, w_m)$. Sentence extraction aims to create a summary from D by selecting a subset of M sentences (where $M < n$). we predict the label of the $j^{th}$ sentence $y_j$ as (0,1). The labels of sentences in the summary set are set as $y = 1$.*

To set the scene of this work, we begin with a brief overview about the self-attention. Given a query vector representation $q$ and an input sequence of tokens $x = [x_1; x_2; \ldots; x_n]$, where $x_i$ denotes the embedded vector of the i-th token); then, the function $f(x_i; q)$ is used to calculate an alignment score between $q$ and each token $x_i$ as the vanilla attention of $q$ to $x_i$ [11]. Self-attention is a special case of attention where the query $q$ stems from the input sequence itself. Therefore, self-attention mechanism can model the dependencies between tokens from the same sequence. The function $f(x_i; x_j)$ is used to compute the dependency of $x_j$ on another token $x_i$, where the query vector $q$ is replaced by the token $x_j$.

The work of Yang *et al.* [8] demonstrates that using the hierarchical attention yields a better document representation, which they used for document classification task. In this work, we propose a new hierarchical structured self-attention architecture modeled by recurrent neural networks based on recent neural extractive summarization approaches [2]–[4]. However, our summarization framework is applicable to all models of sentence extraction using the distributed representation as input.

The proposed model has a hierarchical self-attention structure which reflects the hierarchical structure of the document where words form sentences and sentences form a document. In the new model, there are two level of attention, one at the word-level and the second at the sentence-level. **FIGURE 1** shows the overall architecture of the proposed model that will be explained in the following sections.

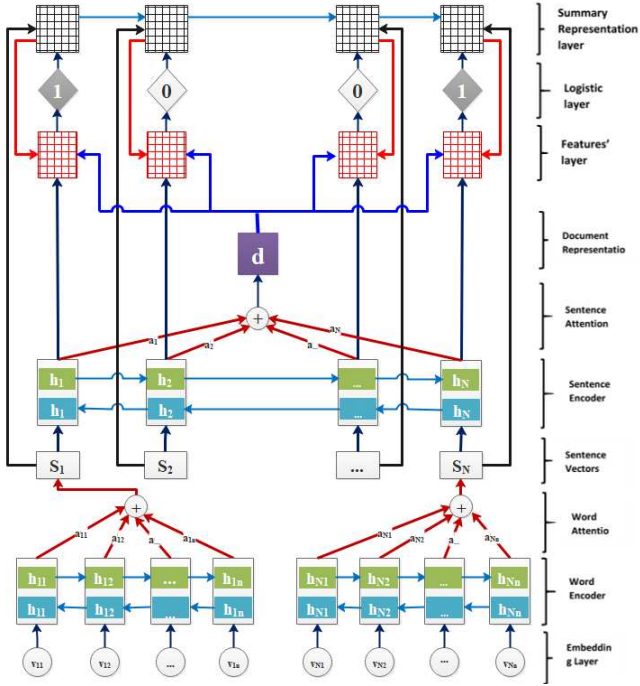

**FIGURE 1.** A Hierarchical Structured Self-attention-based summary-sentence classifier: the first layer is a word-level layer for each sentence. The second layer operates on the sentence-level. After each layer, there is attention layer. The logistic layer is the classification layer which determines the sentence-summary membership, where 1's indicate that the sentence is a summary sentence and 0's determine that the sentence is not

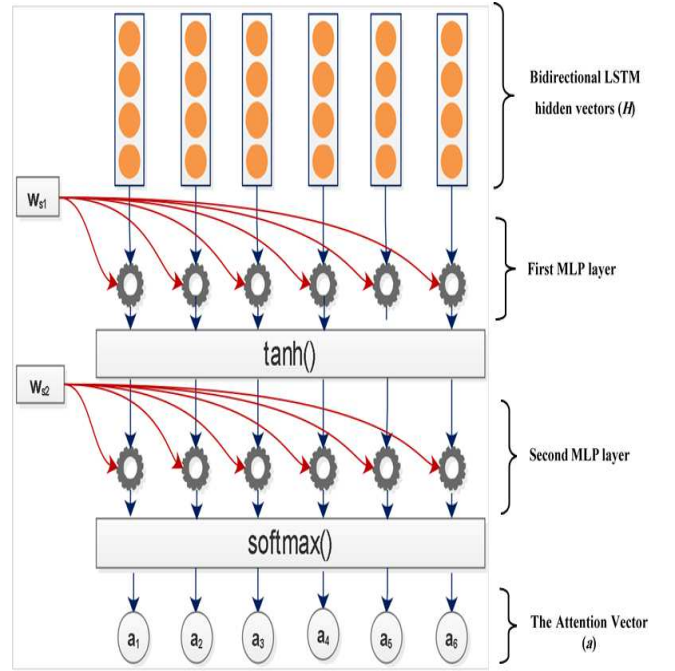

**FIGURE 2.** The Self-Attention Unit. The attention mechanism takes the whole LSTM hidden states H with $\mathbb{R}^{nx2u}$ dimension as an input, and outputs a vector of attention weights, a. Here $w_{s1}$ is a weight matrix with a shape of $\in \mathbb{R}^{kx2u}$. and $w_{s2}$ is a vector of parameters with a size of $\mathbb{R}^k$, where k is a hyperparameter can be set arbitrarily.

### A. WORD ENCODER

Suppose we have a document ($D$) with $n$ sentences and $m$ words, let $D = (s_1, s_2, \ldots, s_n)$ where $s_j (1 \leq j \leq n)$ denotes the $j^{th}$ sentence, and $V = (v_1, v_2, \ldots, v_m)$ where $v_i (1 \leq i \leq m)$ is a vector standing for a $d$ dimensional word embedding for the $i^{th}$ word. In this work, we use a bidirectional LSTM to encode the words in the sentences. The bidirectional LSTM summarizes information from both directions. It contains the forward LSTM which reads the sentence $s_i$ from $v_{i1}$ to $v_{im}$ and a backward LSTM which reads from $v_{im}$ to $v_{i1}$:

$$\overrightarrow{h_t} = \overrightarrow{LSTM}(v_t, \overrightarrow{h_{t-1}}) \quad (1)$$
$$\overleftarrow{h_t} = \overleftarrow{LSTM}(v_t, \overleftarrow{h_{t-1}}) \quad (2)$$

To obtain the hidden state $h_t$ that summarizes the information of the whole sentence centered around $v_{it}$, we concatenate each $\overrightarrow{h_t}$ with $\overleftarrow{h_t}$, as in Equation 3.

$$h_t = [\overrightarrow{h_t}, \overleftarrow{h_t}] \quad (3)$$

Where the number of the hidden units for each unidirectional LSTM is $u$. Let $H_s \in \mathbb{R}^{nx2u}$ denotes the whole LSTM hidden states, as in Equation 4:

$$H_s = (h_1, h_2, \ldots h_n) \quad (4)$$

### B. WORD ATTENTION

The intuition behind the attention mechanism is to pay more or less attention to words according to their contribution to the representation of the sentence meaning. Our objective is to encode a variable length sentence into a fixed size embedding using a self-attention mechanism [7] that takes the whole LSTM hidden states $H_s$ as input and yields a vector of weights, $a_s$, as output, as in Equation 5.

$$a_s = softmax(w_{s_2} tanh(w_{s_1} H_s^T)) \quad (5)$$

Where $w_{s_2} \in \mathbb{R}^{kx2u}$ and $w_{s_1} \in \mathbb{R}^k$ are learnable parameters; $k$ is a hyperparameter can be set arbitrarily. The $softmax()$ is used to normalize the attention weights to sum up to 1.

Given the attention vector $a_s$, we obtain the sentence vector as a weighted sum of the LSTM hidden states weighted by $a_s$, as shown in **FIGURE 2** and Equation 6:

$$s_i = a_s H_s \quad (6)$$

### C. SENTENCE ENCODER

After getting the sentence vector $s_i$, we can get the document representation in the same way. A bidirectional LSTM is used to encode the sentences:

$$\overrightarrow{h_{st}} = \overrightarrow{LSTM}(s_i, \overrightarrow{h_{t-1}}) \quad (7)$$
$$\overleftarrow{h_{st}} = \overleftarrow{LSTM}(s_i, \overleftarrow{h_{t+1}}) \quad (8)$$

Similar to the word encoder, the forward and backward LSTM hidden states are concatenated to get the annotation $h_{st}$, which summarizes the adjacent sentences around the sentence $s_i$ but focus on the $i^{th}$ sentence, as in Equation 9.

$$h_{st} = [\overrightarrow{h_{st}}, \overleftarrow{h_{st}}] \quad (9)$$

Let $u$ denotes the number of the hidden units for each unidirectional LSTM, $N$ is the number of sentences in the $d^{th}$ document, and $H_d$ denotes the whole LSTM hidden states calculated by Equation 10, the dimension of $H_d$ is $\mathbb{R}^{Nx2u}$.

$$H_d = (h_{s1}, h_{s2}, \ldots h_{sN}) \quad (10)$$

### D. SENTENCE ATTENTION

Every sentence in a document contributes differently to the representation of the whole meaning of the document. The self-attention mechanism used in this work takes the whole LSTM hidden states $H_d$ as input and yields a vector of weights, $a_d$, as output, calculated by Equation 11:

$$a_d = softmax(w_{s_2} tanh(w_{s_1} H_d^T)) \quad (11)$$

Where $w_{s_2}$ and $w_{s_1}$ are learnable parameters. The *softmax*() is used to normalize the attention weights to sum up to 1.

Given the attention vector $a_d$, we obtain the document vector as a weighted sum of the LSTM hidden states weighted by $a_d$, as shown in **FIGURE 2**, and Equation 12:

$$d = a_d H_d \quad (12)$$

### E. CLASSIFICATION LAYER

Inspired by an interesting work proposed by Nallapati *et al.* [2], we used a logistic layer that makes a binary decision to determine whether a sentence belongs to the summary or not. The classification decision at the $j^{th}$ sentence depends on the representation of the abstract features, such as the sentence's content richness $C_j$, its salience with respect to the document $M_j$, the novelty of the sentence with respect to the accumulated summary $N_j$ and the positional feature $P_j$. The probability of the sentence belonging to the summary is given by Equation 18:

The information content of the $j^{th}$ sentence in the document is represented by Equation 13:

$$C_j = W_c s_j \quad (13)$$

Equation 14 captures the salience of the sentence with respect to the document:

$$M_j = s_j^T W_s d \quad (14)$$

Equation 15 models the novelty of the sentence with respect to the current state of the summary:

$$N_j = s_j^T W_r \tanh(o_j), \quad (15)$$

where $o_j$ is the summary representation calculated by Equation 17.

The position of the sentence with respect to the document is modeled by Equation 16:

$$P_j = W_p p_j, \quad (16)$$

where $p_j$ is the positional embedding of the sentence calculated by concatenating the embeddings corresponding to the forward and backward position indices of the sentence in the document.

$W_c$, $W_s$, $W_p$, and $W_r$ are automatically learned scalar weights to model the relative importance of various abstract features.

The summary representation, $o_j$, at the sentence $j$ is calculated using Equation 17:

$$o_j = \Sigma_{i=1}^{j-1} s_i P(y_i = 1 | s_i, o_i, d) \quad (17)$$

Where $y_j$ is a binary number determines whether the sentence $j$ is included in the summary or not.

Putting Equations 13, 14, 15 and 16 together, we get the final probability distribution for the sentence label $y_j$, as in Equation 18:

$$P(y_j = 1 | s_j, o_j, d) = \sigma(C_j + M_j - N_j + P_j + b), \quad (18)$$

where $\sigma$ is the sigmoid activation function, and $b$ is the bias term.

Including the summary representation, $o_j$, in the scoring function allows the model to take into account the previously made decisions in terms of determining the summary-sentence membership.

At training phase, the negative log-likelihood of the observed labels is minimized, as in Equation 19:

$$l(W, b) = -\Sigma_{d=1}^{N} \Sigma_{j=1}^{n_d} (y_j^d \log P(y_j^d = 1 | s_j, o_j, d_d) \\ + (1 - y_j^d) \log(1 - P(y_j^d = 1 | s_j, o_j, d_d)), \quad (19)$$

where $y_j^d$ is the binary summary label for the $j^{th}$ sentence in the $d^{th}$ document, $n_d$ is the number of the sentences in the document ($d$), and $N$ is the number of documents.

## III. EXPERIMENTS
### A. DATASETS

The proposed model was evaluated on two well-known datasets, CNN/Daily Mail and DUC 2002. The first dataset was originally built by Hermann *et al.* [12] for question answering task and then re-used for extractive [2], [3] and abstractive text summarization tasks [13], [14]. From the Daily Mail dataset, we used 196,557 documents for training, 12,147 documents for validation and 10,396 documents for testing. In the joint CNN/Daily Mail dataset, there are 286,722 for training, 13,362 for validation and 11,480 for testing. The average number of sentences per document is 28. One of the contributions of [2] is preparing the joint CNN/Daily Mail dataset for the extractive summarization task. In which, they provide sentence-level binary labels for each document, representing the summary-membership of the sentences. We refer the reader to that paper for a detailed description.

The second dataset is DUC 2002 used as an out-of-domain test set. It contains 567 news articles belonging to 59 different clusters of various news topics, and the corresponding 100-word manual summaries generated for each of these documents (single-document summarization), or the 100-word summaries generated for each of the 59 document clusters formed on the same dataset (multi-document summarization). In this work, we used the single-document summarization task.

### B. BASELINES

There are so many approaches to the text summarization problem; for comparison, we choose the ones that are comparable to our work on the two datasets as follows:

- Leading sentences (Lead-3): which simply produces the first three sentences of the document as a summary. This model serves as a baseline on the two datasets, CNN/Daily Mail and DUC 2002.
- Recurrent Neural Network based model (SummaRuN-Ner) proposed by Nallapati *et al.* [2], mentioned in section 4, is used as a baseline on the two datasets.
- In addition, the extractive model proposed by Cheng and Lapata [3] was used as a baseline on the two datasets.
- On CNN/Daily Mail dataset, we used the reinforced abstractive summarization model proposed by Paulus *et al.* [13] and a pointer-Generator based Network by See *et al.* [15] as abstractive baselines.
- We also used TGRAPH [16], a graph based approaches, and URANK [17] as baselines on DUC 2002 as they achieve high performance on this dataset.

### C. SETTINGS

We got the word embedding initialization by training word2vec [2] on the CNN/Daily Mail dataset. The validation set was used to tune the hyperparameters. The word embedding dimension was set to 100 and the model hidden state size to 200. The concatenation of forward and backward LSTMs gives us a dimension of 400 for both word encoder and sentence encoder. The word and sentence attention context vectors also have a dimension of 400. The vocabulary size was limited to 150k. We set the maximum sentence length to 50 words and the maximum number of sentences per document to 100. At training time, the batch size was 64, and adadelta [18] was used to train the model and the gradient clipping to regularize it. At test time, we sorted the output probabilities for the sentence-summary membership and then pick the sentences with the top probabilities until we exceed the compression rate.

### D. EVALUATION

In this work, the ROUGE (Recall-Oriented Understudy for Gisty Evaluation) metrics [19] are used for the automatic evaluation of the generated summaries. ROUGE metrics are based on the comparison of n-grams between the summary to be evaluated and one or several human written reference

**TABLE 1.** The performance comparison of the proposed models with respect to the baselines on DUC 2002.

| Model | DUC 2002 | | |
| --- | --- | --- | --- |
| | ROUGE-1 | ROUGE-2 | ROUGE-L |
| URANK [17] | 0.485 | 0.215 | - |
| TGRAPH [16] | 0.481 | 0.243 | - |
| LEAD-3 | 0.436 | 0.210 | 0.402 |
| Cheng et al '16 [3] | 0.474 | 0.230 | 0.435 |
| SummaRuNNer [2] | 0.474 | 0.221 | 0.420 |
| Ours (HSSAS) | **0.521** | **0.245** | **0.488** |

summaries, as in Equation 20.

$$\text{ROUGE}_N = \frac{\Sigma_{S\in\{reference\ Summaries\}} \Sigma_{gram_n \in S} Countmatch(gram_n)}{\Sigma_{S\in\{reference\ Summaries\}} \Sigma_{gram_n \in S}(gram_n)} \quad (20)$$

*Remark 1:* To ensure that the recall-only evaluation will be unbiased to length, we use the "-l 75" options in ROUGE to truncate longer summaries in DUC 2002.

*Remark 2:* It is noticed that all the baselines use full-length F1 as an evaluation metric on the entire CNN/DailyMail since the neural abstractive models learn when to stop generating word for the summary. To ensure a fair comparison, we apply the same metric.

### E. EXPERIMENTAL RESULTS

The ROUGE Toolkit[1] and the pyrouge package[2] are used to evaluate the performance of the proposed model. ROUGE-1, ROUGE-2, and ROUGE-L were applied with the settings that mentioned in Remark 1 and Remark 2. We compared our model with several extractive and abstractive baselines, mentioned in section 3.2. The model and the baselines are evaluated on the two datasets, CNN/Daily Mail and DUC 2002. The output of the evaluation was compared to the human-generated summaries in these datasets. The evaluation results, shown in **Table 1**, **TABLE 2**, **FIGURE 3**, and **FIGURE 4**, assert that the proposed model achieves promising results. From the obtained results, we can make the following observations:

- As shown in **Table 1** and **FIGURE 3**, the obtained results for ROUGE-1, ROUGE-2, and ROUGE-L indicate that our proposed method, HSSAS, performs the best for all ROUGE metrics used in this experiment on DUC 2002 dataset. This asserts that using hierarchical self-attention leads to better sentence and document representations and enhances the abstract features that can be used to yield state-of-the-art performance on the text summarization task.
- In the case of CNN/Daily Mail, **TABLE 2** and **FIGURE 4**, the results assert that the proposed models, HSSAS, outperforms all the baselines in the term of almost all ROUGE metrics used in this experiment.

---

[1] http://www.berouge.com/Pages/default.aspx ROUGE-1.5.5 with options: -n 2 -m -u -c 95 -r 1000 -f A -p 0.5 -t 0

[2] https://pypi.python.org/pypi/pyrouge/0.1.3

TABLE 2. The performance comparison of the proposed models with respect to the baselines on CNN/Daily Mail using full-length F1 variant of ROUGE.

|  | CNN/Daily Mail | | |
| --- | --- | --- | --- |
| Model | ROUGE-1 | ROUGE-2 | ROUGE-L |
| LEAD-3 | 0.392 | 0.157 | 0.355 |
| Cheng et al '16 [3] | 0.354 | 0.133 | 0.326 |
| SummaRuNNer [2] | 0.399 | 0.163 | 0.351 |
| Pointer-generator + coverage [15] | 0.395 | 0.173 | 0.364 |
| RL, with intra-attention [13] | 0.416 | 0.157 | **0.391** |
| Ours (HSSAS) | **0.423** | **0.178** | 0.376 |

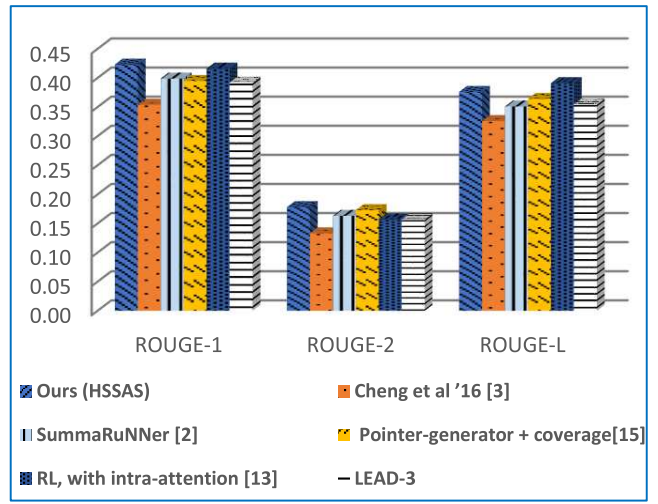

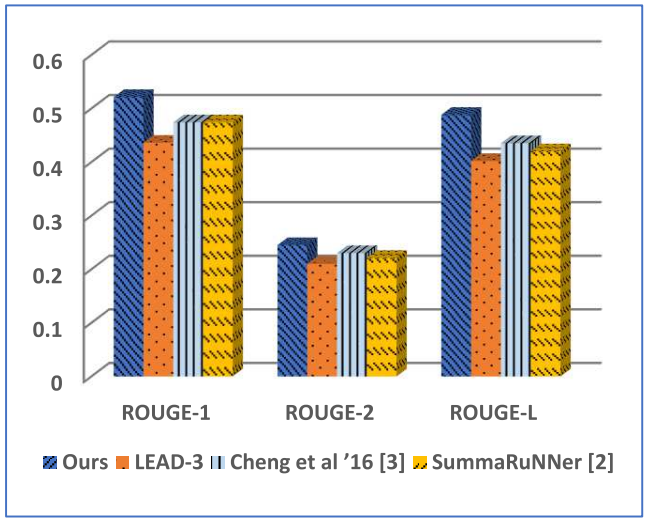

FIGURE 3. The performance comparison of the proposed model with respect to the baselines on DUC 2002.

FIGURE 4. The performance comparison of the proposed model with respect to the baselines on CNN/Daily Mail dataset using full-length F1 variant of ROUGE.

- In news articles, it is usual for the important information to be put in the beginning of the article. This justifies the good ROUGE results of the LEAD-3 baseline in DUC 2002 which makes it hard to be beaten; however, our model has performed better.
- In the context of abstractive based models, while ROUGE measures the n-gram overlap between the generated summary and a reference one, summaries with high ROUGE scores are not necessarily the more readable ones. One potential issue of generative summarization models is that optimizing for a specific discrete metric like ROUGE does not guarantee an increase in quality and readability of the generated summary [13], [20]. This may justify the competitive ROUGE scores of the abstractive baselines used in this work.
- Another issue related to the ROUGE metric is that the reliability of ROUGE increased by the number of the reference summaries per document. This inflexibility of ROUGE makes the Rouge scores on the datasets that has one reference summary per document much lower compared to the ones that have multiple reference summaries [15], [19].
- Finally, our proposed model, HSSAS, obtained good results competing with the state-of-the-art methods.

## IV. RELATED WORK

Summarization systems fall into two main categories, extractive and abstractive. Extractive summarization models generate the summary by extracting some key subset of the content for the original document in a way that this subset contains the core information. By contrast, abstractive summarization models are more sophisticated and more complex since they leverage the language semantics to create representations. They use different words to describe the contents of the original documents rather than extracting the original ones [13], [21].

Since it is comparatively harder to build abstraction-based summarizers, most of the previously proposed models focus on the extraction-based models. Recently, neural network methods have been being used for extractive summarization. For example, a recursive autoencoder based model proposed by Kragebäck et al. [22] to summarize documents on the Opinosis dataset [23]. For multi-document extractive summarization task, Yin and Pei [24] have used Convolutional Neural Networks (CNN) to project sentences to a continuous vector space and then they used the sentence 'diverseness' and 'prestige' to minimizing the cost function. A query-focused model for multi-document summarization was proposed by Cao et al. [25]. In which, they addressed the problem using query-attention-weighted CNNs. Another extractive summarization approach has been proposed by Nallapati et al. [26]. They used an RNN based classifier that sequentially labels the sentences with binary labels 0/1 for their membership in the summary. The score for each sentence is computed by explicitly modeling abstract features such as content richness, salience with respect to the document.

For extractive query-oriented single-document summarization, Yousefi-Azar and Hamey [27] used a deep autoencoder to compute a feature space from the term-frequency (*tf*) input. They developed a local word representation in which each vocabulary is designed to build the input representation for sentences in the document. Then, a random noise is added to the word representation vector, affecting both the input and output of the auto-encoder.

A recent work proposed by See *et al.* [15] in which they augmented the standard sequence-to-sequence attentional model in two orthogonal ways. In first way, they used a hybrid pointer-generator network that can copy words from the source text via pointing. In the second one, they used the coverage to keep track of what has been summarized so far. Another study carried out by Cao *et al.* [25] tried to learn the distributed representations for sentences by applying an attention mechanism, which used to learn query relevance ranking and sentence saliency ranking simultaneously. Another extractive summarization model was proposed by Cheng and Lapata [3] in which they have treated the single document summarization as a sequence labeling task using a document encoder and attention-based extractor. They applied the attention directly to extract sentences and words for the summary.

The most similar work to ours is the one proposed by Nallapati *et al.* [2]. They used a recurrent neural network (RNN) based model for extractive summarization applied to the CNN/Daily Mail corpus. In which, they treated extractive summarization as a sequence classification problem. They used neural networks for the sentential extractive summarization of single documents. In their model, each sentence is visited sequentially as it appears in the original document and a binary decision is taken to determine whether the sentence should be included in the summary or not. It is worth mentioning that they did not use any attention mechanism. Different from their approach, our model uses the structured self-attentive mechanism that has the capability to guide the sentence and document representations.

The recent advancement of the generative neural models for text makes the abstractive summarization techniques increasingly popular. In 2015, Rush *et al.* [28] published an encoder-decoder model, in which the encoder is a convolutional network and the decoder is a feedforward neural network language model. They enhanced the convolutional encoder by integrate it with attention model. Then they used the trained neural network as a feature to a log-linear model. As the convolutional encoder need a fix number of features, they used a bag of n-grams model. That means they ignore the overall sequence order while generating the hidden representation. They only used the first sentence of each news article to generate its title. Another recent abstractive model was proposed by Paulus *et al.* [13]. In which, they combined the standard supervised word prediction with reinforcement learning (RL).

Despite the popularity of abstractive techniques, extractive techniques are still attractive as they are less expensive, less complex and most of the time, they can generate grammatically and semantically correct summaries. Moreover, the performance of RNN-based encoder-decoder models for abstractive summarization is quite good for short input and output sequences, but for longer documents and summaries, these models often struggle from serious problems such as repetition, unreadability and incoherence.

As we mentioned earlier in the paragraph before the last one in section 1, our work differs from the previous ones by it is capability to capture the hierarchical structure of the documents. Moreover, it uses the hierarchical structured self-attention to deliver a better embedding representation for sentences and documents. The attention mechanism that we use in this work puts more focus on the semantics of the whole sentence that each word contributes to rather than just focusing on the relations between words like the previous attention-based models.

## V. CONCLUSION AND FUTURE WORK

The proposed model is another way of utilizing the attention mechanism to create a sentence and document embeddings. The experimental results of the proposed model assert that those embeddings deliver a better representation which in turn enhances the document summarization task and outperforms the state-of-the-art models on the same datasets. This work is different from the previous work in the sense of three points. First, it uses the hierarchical attention that mirror the document structure. Second, it uses the structured self-attention, which creates a very good embedding. Third, the abstract features are weighted and automatically learned during the learning process taking in consideration the previously classified sentences. We believe that combining the reinforcement learning with sequence-to-sequence training objective is an interesting direction for further research. Another research effort should be directed toward proposing another evaluation metric beside ROUGE metric to optimize on summarization model especially for long sequences.

### ACKNOWLEDGEMENTS
We are grateful to the support of the National Natural Science Foundation of China (Grant No. 61379109, M1321007) and Science and Technology Plan of Hunan Province (Grant No. 2014GK2018 ,2016JC2011).